\pdfoutput=1
\documentclass[11pt,a4paper]{article}
\usepackage[hyperref]{ranlp2021}
\usepackage{times}
\usepackage{latexsym}
\usepackage{comment}
\usepackage{todonotes}
\usepackage{booktabs}
\usepackage{adjustbox}
\usepackage{multirow}
\usepackage{amsmath}

\usepackage{microtype}

\aclfinalcopy 

\title{Multilingual Coreference Resolution with Harmonized Annotations}

\author{Ondřej Pražák \and Miloslav Konopík \and Jakub Sido\\[0.5em]
\tt{\{ondfa,sidoj\}@ntis.zcu.cz}\\
\tt{konopik@kiv.zcu.cz}\\[0.5em]
Department of Computer Science and Engineering, \\
NTIS -- New Technologies for the Information Society, \\
Faculty of Applied Sciences, University of West Bohemia, Technick\'a 8, 306 14 Plze\v{n} \\
Czech Republic
}

\date{}

\begin{document}

\maketitle
\begin{abstract}
In this paper, we present coreference resolution experiments with a newly created multilingual corpus CorefUD \cite{Nedoluzhko2021}. We focus on the following languages: Czech, Russian, Polish, German, Spanish, and Catalan. In addition to monolingual experiments, we combine the training data in multilingual experiments and train two joined models -- for Slavic languages and for all the languages together. We rely on an end-to-end deep learning model that we slightly adapted for the CorefUD corpus. Our results show that we can profit from harmonized annotations, and using joined models helps significantly for the languages with smaller training data.
\end{abstract}

\section{Introduction}

Coreference resolution is the task of finding language expressions that refer to the same real-world entity (antecedent) of a given text. Sometimes the corefering expressions can come from a single sentence. However, the expressions can be one or more sentences apart as well. It is necessary to see the whole document in some hard cases to judge whether two expressions are corefering adequately. This task can be divided into two subtasks. Identifying entity mentions, and grouping the mentions together according to the real-world entity they refer to. The task of coreference resolution is closely related to anaphora resolution -- see \cite{Sukthanker2020} to compare these two tasks. 

One of the challenging difficulties of coreference resolution lay in linguistically complicated annotations. 
Some examples of linguistic complications are \emph{split antecedents} (a mention refer to more than one real-word entities), \emph{near identity} relations, \emph{anaphoric} and \emph{cataphoric}
relations, etc \cite{Nedoluzhko2021}. 

In this paper, we rely on a CorefUD corpus \cite{Nedoluzhko2021} of harmonized annotations. This corpus enables us to battle linguistic complications since it simply presents corefering mentions in clusters. 
Since the corpus is compiled from 11 different corpora in 8 different languages, we can conduct multilingual experiments in this work. Our research goal is to evaluate whether the harmonized annotations open the possibility to obtain some performance gain by joint learning on multiple languages. We aim to compare the harmonized annotations with the original corpora as well.

\section{Related Work}

In agreement with many other NLP tasks, deep learning models prevail in the coreference resolution task. \citet{lee-etal-2017-end} were first to introduce the end-to-end approach that many following papers adopted (they obtained an average of 67.2 of F1 score). The task experienced a big leap in performance with the introduction of large pre-trained models. BERT based models deliver the best results; \citet{kantor-globerson-2019-coreference} F1 76.6, and \citet{joshi-etal-2019-bert} F1 76.9. \citet{joshi-etal-2020-spanbert} came up with a new pretraining task focused on better span representations. Their model called SpanBERT brings additional improvements in the coreference resolution task (F1 79.6). \citet{xu-choi-2020-revealing} question the importance of modeling higher-order inference (HOI). They show that with advanced encoders, HOI has only a minor effect on the performance of models.

Research is significantly less evolved for other languages than English. However, some notable experiments were published in recent years. \citet{recasens-etal-2010-semeval} describe multilingual experiments (for English, Catalan and Spanish, Dutch, German and Italian) similarly to our paper. However, the annotations were not harmonized as in our case. Therefore, they provide no experiments with joint training. 

Other cross-lingual experiments include Portuguese by learning from Spanish \cite{Cruz-etal-2018}; Spanish and Chinese relying on an English corpus \cite{kundu-etal-2018-neural}; and Basque based on an English corpus as well \cite{urbizu-etal-2019-deep}. All these approaches employ neural networks, and they transfer the model via cross-lingual word embeddings. 

Treex CR \cite{Novak17-TreexCR} is a coreference resolution module in the Treex NLP framework\footnote{https://github.com/ufal/treex}. It produces an advanced syntactic analysis with semantic features that the tool uses to find coreference relations -- offers models for Czech and English. 
Other non-English experiments include Polish \cite{niton-etal-2018-deep}, Russian \cite{Sboev-2020}, and German \cite{Srivastava-2018}.

\section{Dataset}
For our experiments, we use the harmonized multilingual coreference dataset \textit{CorefUD} \cite{Nedoluzhko2021}. The dataset was created by converting 17 existing datasets for 11 different languages into a common format on the top of universal syntactic annotations -- Universal Dependencies. For coreference representation, a cluster-based approach was selected instead of the link-based approach. It is simpler and moreover the most frequently used dataset for English -- OntoNotes adopt this approach too. In a cluster-based approach, every mention belongs to one cluster, represented by a unique ID. In a link-based approach, coreferences are expressed by the links between corefering mentions.  In the link-based approach, coreference structures form a chain, but there are more complex coreference structures in some cases \cite{Nedoluzhko2021}. Datasets that use the link-based approach were converted to cluster-based at the cost of some information loss. 

There are some notable differences between the datasets. One of the most prominent ones is the presence of singletons. Singletons are clusters that contain only one mention. Singletons are not present in any coreference relation. However, they are annotated as mentions in all datasets. Discontinued mentions represent another notable difference. A discontinuous mention consists of a sequence of words that is interrupted at least once with some words that do not belong to the mention. Such mentions can cause problems to models that assume mentions to be continuous (such as our model).

Table \ref{tab:data} shows the statistics of the datasets including the above-mentioned differences. More detailed analysis can be found in \citet{Nedoluzhko2021}.

\begin{table*}[ht!]
\catcode`\-=12
\begin{adjustbox}{width=\linewidth,center}
    \centering
    \begin{tabular}{lcccccccccc}
    \toprule
        \multirow{2}{*}{CorefUD dataset} & \multicolumn{6}{c}{Total size} & & \multicolumn{3}{c}{Division [\% of words]}  \\
        \cline{2-7}
        \cline{9-11}
         & docs & sents & words & empty & singletons & discont. & & train & dev & test \\
         \hline
         Catalan-AnCora & 1550 & 16,678 & 488,379 &6,377 & 74.6\% & 0\% & & 78.6 & 10.7 & 10.8 \\
         Czech-PDT & 3165 & 49,428 & 834,721 & 33,086 & 35.3\% & 3.1\% & & 78.3 & 10.6 & 11.1 \\
         German-PotsdamCC & 176 & 2,238 & 33,222 & 0 & 76.5\% & 6.3\% & & 80.3 & 10.2 & 9.5 \\
         Polish-PCC & 1828 & 35,874 & 538,891 & 864 & 82.6\% & 1.0\% & & 80.1 & 10.0& 9.9 \\
         Russian-RuCor & 181 & 9,035 & 156,636 & 0 & 2.5\% & 0.5\% & & 78.9 & 13.5 & 7.6 \\
         Spanish-AnCora & 1635 & 17,662 & 517,258 & 8,111 & 73.4\% & 0\% & & 80.9 & 9.5 & 9.6 \\
         \bottomrule
    \end{tabular}
    \end{adjustbox}
    \caption{Basic dataset statistics including train/dev/test split of CorefUD data sets. Column \textit{discont.} shows the percentage of discontinuous mentions. Taken from \citet{Nedoluzhko2021}.}
    \label{tab:data}
\end{table*}

\section{Model} \label{sec:model}

We use the basic end-to-end model from \citet{xu-choi-2020-revealing} with no higher-order inference (HOI), so it is the same model as it was proposed by \citet{lee-etal-2017-end}.

In the model, we start by modeling the probability $P(y_i|D)$ of a mention $i$ corefering with the antecedent $y_i$ in a document $D$. Since the model adopts the end-to-end approach, the mentions are identified together with the coreference links. We consider every continuous sequence of words as a mention $i$. Therefore, we work with $N = \frac{T(T+1)}{2}$ possible mentions, where $T$ is the number of words in a document $D$.

We model the score of a mention $i$ corefering with an antecedent $y_i$ as a combination of two types scores $s_m(i)$ and $s_a(i,y_i)$. The $s_m$ is a score of a sequence of words (spans) $i$ being a mention. The $s_a(i,y_i)$ score is the score of a span $y_i$ being an antecedent of span $i$. The scores are combined as a sum of $s_m(i)$, $s_m(y_i)$ and $s_a(i,y_i)$ as follows:

\begin{equation}
    s(i, y_i) = 
    \begin{cases}
                0 & y_i = \epsilon \\
                s_m(i) + s_m(y_i) + s_a(i,y_i) & y_i \ne \epsilon\\  
    \end{cases},
\end{equation}\label{eq:scoring}
where $\epsilon$ is an empty antecedent. Both scores $s_m(i)$ and $s_a(i,y_i)$ are estimated with a feed-forward neural network over the BERT-based encoder. In our experiments we use two encoders -- multilingual BERT \cite{bertpaper} and Slavic BERT \cite{arkhipov2019tuning-SlavicBert}.

The probability of an antecedent $y_i$ can be expressed as the $softmax$ normalization over all possible antecedents $y' \in Y(i)$  for a mention $i$:

\begin{equation}
    P(y_i|D) = \frac{\exp(s(i, y_i)}{\sum_{y' \in Y(i)}\exp(s(i, y')} 
\end{equation}

The formula for all antecedents uses a product of multinomials of all individual antecedents:

\begin{equation}
    P(y_1, ..., y_N|D) = \prod_{i=1}^N P(y_i|D)
\end{equation}

In the training phase, we maximize the marginal log-likelihood of all correct antecedents:

\begin{equation}
    J(D) = \log \prod_{i=1}^N \sum_{\hat{y} \in Y(i) \cap \texttt{GOLD}(i)}P(\hat{y})
\end{equation}\label{eq:loss}

where GOLD($i$) is the set of spans in the training data that are  antecedents.
\section{Experiments}

First, we perform monolingual experiments with the model described in Section \ref{sec:model} on several largest datasets from \textit{CorefUD}. Namely Czech, Russian, Polish, Spanish, Catalan, and German-PotsdamCC. The employed datasets are summarized in Table \ref{tab:data} along with some basic statistics. The datasets are split to train, dev, and test, but the test datasets are not publicly available. Therefore, we use the original dev datasets as test datasets, and we create new dev datasets by taking 10\% of the training parts. We tune the hyperparameters and perform early stopping on the development parts.  

As the next step, we perform multilingual experiments, where we train one model for all the Slavic languages (Czech, Russian, and Polish) and another model for all the languages (Czech, Russian, Polish, German, Spanish, and Catalan). Multilingual results in comparison with the monolingual ones are shown in Table \ref{tab:results-all}. 

The results in Table \ref{tab:results-all} are influenced by the presence of singletons in the datasets. Particularly, Polish, German, Spanish, and Catalan contain a large portion of singletons, which negatively impacts the results. Since our employed model cannot model singletons, we have removed them from the test datasets. We show the results on filtered datasets in Table \ref{tab:results-all-filtered}. Singletons are not important for coreference resolution since they form no coreference relation. However, they can be important in the training phase, if the model can use them for mention recognition.  

We report the average F1 measure from the official scoring script\footnote{https://github.com/conll/reference-coreference-scorers} for the coreference resolution task in CoNLL. The metric is computed as the average of $MUC$, $B^3$ and $CLEAF_{4}$. Definition of these metric can be found in \citet{pradhan-EtAl:2014:P14-2}. The F1 scores are reported with 95\% confidence intervals measured from 5 runs.  We use the same set of hyperparameters for all the languages and for all models. We train the models for approximately for 100k steps. We employ the Adam optimizer with the learning rate of 0.00001 for BERT layers and 0.0002 for other layers.

\begin{table*}[ht!]
\catcode`\-=12
\begin{adjustbox}{width=\linewidth,center}
\begin{tabular}{lcccccc}
\toprule
                  & czech  & russian & polish & german & spanish & catalan \\
\midrule
Mono-mBERT             & $58.883 \pm 0.204$ & $62.665 \pm 1.028$  & $42.411 \pm 0.303$ & $39.958 \pm 0.775$ & $49.654 \pm 0.118$  & $47.962 \pm 0.302$  \\
Mono-SlavicBert        & $\textbf{60.283} \pm \textbf{0.013}$ & $62.097 \pm 1.153$  & $43.234 \pm 0.114$ &     -   &      -   &   -      \\
Slavic-mBERT      & $58.734 \pm 0.198$ & $\textbf{66.762} \pm \textbf{0.495}$  & $44.091 \pm 0.413$ &   -     &       -  &     -    \\
Slavic-SlavicBERT & $60.096 \pm 0.103$ & $64.414 \pm 0.750$  & $\textbf{44.943} \pm \textbf{0.110}$ &   -     &      -   &     -    \\
Joined-mBERT      & $58.990 \pm 0.304$ & $65.243 \pm 0.942$  & $44.346 \pm 0.342$ & $\textbf{46.098} \pm \textbf{0.641}$ & $\textbf{51.192} \pm \textbf{0.221}$  & $\textbf{49.881} \pm \textbf{0.126}$  \\
\bottomrule
\end{tabular}
\end{adjustbox}
\caption{Overall results of F1 averages obtained from the official scoring script.}
\label{tab:results-all}
\end{table*}

\begin{table*}[]
\centering
\begin{tabular}{lllllll}
\toprule
                  & czech  & russian & polish & german & spanish & catalan \\
\midrule
Joined-mBERT             & +0.107  & +1.926   & \textbf{+1.935}  & \textbf{+6.140}   & \textbf{+1.538}   & \textbf{+1.919}   \\
Slavic-mBERT      & -0.149 & \textbf{+3.445}   & \textbf{+1.680}   &    -    &    -     &   -      \\
Slavic-SlavicBERT & -0.187 & \textbf{+2.317}   & \textbf{+1.709}  &    -    &     -    &     -    \\
\bottomrule
\end{tabular}

\caption{F1 gains of multilingual models over the same monolingual ones. Bold numbers indicate that the difference is bigger than the width of confidence interval. Table depicts absolute differences.}
\label{tab:differences}
\end{table*}

\begin{table*}[ht!]
\catcode`\-=12
\begin{adjustbox}{width=\linewidth,center}
\begin{tabular}{lcccccc}
\toprule
                  & czech  & russian & polish & german & spanish & catalan \\
\midrule
Mono-mBERT             & $64.383 \pm 0.153$ & $63.135 \pm 0.521$  & $60.247 \pm 0.242$ & $52.541 \pm 1.183$ & $67.88 \pm 0.543$  & $64.394 \pm 0.685$  \\
Mono-SlavicBert        & $\textbf{65.835} \pm \textbf{0.141}$ & $63.453 \pm 0.615$  & $61.726 \pm 0.395$ &     -   &      -   &   -      \\
Slavic-mBERT      & $63.980 \pm 0.211$ & $\textbf{66.794} \pm \textbf{1.105}$  & $61.584 \pm 0.396$ &   -     &       -  &     -    \\
Slavic-SlavicBERT & $65.443 \pm 0.231$ & $64.192 \pm 0.475$  & $\textbf{62.883} \pm \textbf{0.068}$ &   -     &      -   &     -    \\
Joined-mBERT      & $64.176 \pm 0.120$ & $65.618 \pm 0.314$  & $61.959 \pm 0.431$ & $\textbf{61.439} \pm \textbf{1.216}$ & $\textbf{68.9825} \pm \textbf{0.209}$  & $\textbf{66.456} \pm \textbf{0.092}$  \\
\bottomrule
\end{tabular}
\end{adjustbox}
\caption{Overall results of F1 averages obtained from the official scoring script after singleton removal.}
\label{tab:results-all-filtered}
\end{table*}

\section{Discussion}

From the result (See Table \ref{tab:differences}), we can see that joined multilingual models helps all the languages except for Czech -- which does make sense because the Czech dataset is the largest one in the CorefUD corpus.

For the smallest datasets (Russian and German), multilingual models outperform the monolingual ones by a large margin (cca 2 -- 6 \% F1). We can see that at least in small training datasets, using joined models definitely helps, and we can profit from the harmonized coreference annotations.  It is also worth noticing that the confidence intervals for these datasets are significantly wider than for other datasets.

After the singleton filtering the employed model achieves good results for all languages tested.


For German, there are 6.3\% of discontinuous entity mentions. The model iterates over all possible continuous spans. Therefore, it is not able to identify discontinuous mentions at all. For Geraman, the effect of singletons and discontinuous mentions combines.


\section{Future Work}

Currently, we experimented only on a subset of languages available in \textit{CorefUD}. This was caused mainly by the resource-exhaustive training. We need 32GB graphic cards to capture long-enough contexts. We plan to experiment with the rest of the languages in the future.


Additionally, it would be interesting to explore the possibilities of zero-shot cross-lingual transfer in \textit{CorefUD}, where we will not use the training data for the target language at all. 

\section{Conclusion}

We performed pilot experiments to evaluate inter-language transferability of the models based on the \textit{CorefUD} dataset. To do so, we used an end-to-end coreference resolution model based on BERT-like models. Our experiments show that learning from multiple languages in CorefUD annotation scheme helps significantly especially for languages with smaller training data (like Russian and German data in \textit{CorefUD}).

\section*{Acknowledgments}
This work has been partly supported from Grant No. SGS-2019-018 Processing of heterogeneous data and its specialized applications. Computational resources were supplied by the project "e-Infrastruktura CZ" (e-INFRA LM2018140) provided within the program Projects of Large Research, Development and Innovations Infrastructures.

\bibliographystyle{acl_natbib}
\bibliography{anthology,ranlp2021}


\end{document}